\documentclass[conference,letterpaper]{IEEEtran}
\IEEEoverridecommandlockouts
\usepackage{cite}
\usepackage{amsmath,amssymb,amsfonts}
\usepackage{algorithmic}
\usepackage{graphicx}
\usepackage{textcomp}
\usepackage{xcolor}
\usepackage{subfigure}
\usepackage{caption}

\def\BibTeX{{\rm B\kern-.05em{\sc i\kern-.025em b}\kern-.08em
    T\kern-.1667em\lower.7ex\hbox{E}\kern-.125emX}}

\begin{document}

\title{PAtt: A Pattern Attention Network for ETA Prediction Using Historical Speed Profiles}

\author{
\IEEEauthorblockN{ByeoungDo Kim$^{*}$, JunYeop Na, Kyungwook Tak, JunTae Kim, DongHyeon Kim, Duckky Kim}
\IEEEauthorblockA{
% Naver Corporation\\
\textit{Naver Corporation}\\
\{byeoungdo.kim, junyeop.na, kyungwook.tak, jun\_tae.kim, donghyeon.kim67, duckky.kim\}@navercorp.com}
% \and
% \IEEEauthorblockN{2\textsuperscript{nd} Given Name Surname}
% \IEEEauthorblockA{\textit{dept. name of organization (of Aff.)} \\
% \textit{name of organization (of Aff.)}\\
% City, Country \\
% email address or ORCID}
% \and
% \IEEEauthorblockN{3\textsuperscript{rd} Given Name Surname}
% \IEEEauthorblockA{\textit{dept. name of organization (of Aff.)} \\
% \textit{name of organization (of Aff.)}\\
% City, Country \\
% email address or ORCID}
% \and
% \IEEEauthorblockN{4\textsuperscript{th} Given Name Surname}
% \IEEEauthorblockA{\textit{dept. name of organization (of Aff.)} \\
% \textit{name of organization (of Aff.)}\\
% City, Country \\
% email address or ORCID}
% \and
% \IEEEauthorblockN{5\textsuperscript{th} Given Name Surname}
% \IEEEauthorblockA{\textit{dept. name of organization (of Aff.)} \\
% \textit{name of organization (of Aff.)}\\
% City, Country \\
% email address or ORCID}
% \and
% \IEEEauthorblockN{6\textsuperscript{th} Given Name Surname}
% \IEEEauthorblockA{\textit{dept. name of organization (of Aff.)} \\
% \textit{name of organization (of Aff.)}\\
% City, Country \\
% email address or ORCID}
\thanks{$^{*}$Corresponding author}
}

\maketitle

\begin{abstract}
In this paper, we propose an ETA model (Estimated Time of Arrival) that leverages an attention mechanism over historical road speed patterns.
As autonomous driving and intelligent transportation systems become increasingly prevalent, the need for accurate and reliable ETA estimation has grown, playing a vital role in navigation, mobility planning, and traffic management.
However, predicting ETA remains a challenging task due to the dynamic and complex nature of traffic flow.
Traditional methods often combine real-time and historical traffic data in simplistic ways, or rely on complex rule-based computations.
While recent deep learning models have shown potential, they often require high computational costs and do not effectively capture the spatio-temporal patterns crucial for ETA prediction.
ETA prediction inherently involves spatio-temporal causality, and our proposed model addresses this by leveraging attention mechanisms to extract and utilize temporal features accumulated at each spatio-temporal point along a route.
This architecture enables efficient and accurate ETA estimation while keeping the model lightweight and scalable.
We validate our approach using real-world driving datasets and demonstrate that our approach outperforms existing baselines by effectively integrating road characteristics, real-time traffic conditions, and historical speed patterns in a task-aware manner.
\end{abstract}


\section{Introduction}

With the widespread adoption of various transportation means, the advancement of intelligent systems, and the development of autonomous driving, current driving practices have significantly changed.
Unlike the past, drivers today often follow predefined routes set by navigation systems. These routes are not only focused on reaching the destination but also consider various factors such as the shortest travel time, the most comfortable route, fuel efficiency, and optimization based on traffic conditions.
As a result, accurate ETA (Estimated Time of Arrival) goes beyond simply predicting arrival times. It plays a crucial role in decision-making for individual drivers, as well as in logistics and transportation scheduling and optimization.\cite{usage1, usage2, usage3, usage4, usage5, usage6, usage7, usage8}
ETA is a key factor in efficient resource allocation, cost reduction, and improving customer satisfaction. However, improving the accuracy of ETA is a challenging task since it is influenced by various traffic factors such as traffic volume, driving speed, and road characteristics.
It is essential to not only consider current traffic conditions but also predict future traffic conditions and incorporate those predictions into ETA calculations. Therefore, ETA prediction methods that are solely based on distance or current traffic conditions are insufficient for providing accurate estimates. More precise ETA predictions require advanced techniques that combine and analyze both past and present traffic data and predict future traffic conditions.

Given the significance and complexity of the problem, ETA prediction has long been regarded as a core challenge within the domain of Intelligent Transportation Systems (ITS), and numerous methods have been proposed to address this challenge.
Previous studies on ETA prediction have explored a variety of approaches, ranging from historical pattern-based methods to deep learning models. Early methods focused on matching past trajectories or applying statistical models based on time and location features \cite{sbtte,ttepst, bus_arrival4, bus_arrival3, bus_arrival2}.
With the rise of data availability and computational power, recent works have leveraged neural networks, including RNNs, CNNs, and GNNs, to capture complex spatiotemporal dependencies across road networks \cite{deepreta, deeptravel, dcrnn, stgcn, stann, constgat, dueta, gnneta, compacteta, metaer, eep, gbtte, ssml}. These models emphasize real-time traffic dynamics, spatial correlations among road segments, and contextual embeddings. However, challenges remain in terms of scalability, interpretability, and robustness under real-world dynamic traffic conditions.
Some models emphasize real-time speed updates and traffic propagation between links, while others integrate external factors such as weather or events. Despite their advancements, many approaches struggle with generalization, scalability, or real-world applicability, highlighting the need for more interpretable and efficient models.

In this paper, we propose an ETA prediction model based on pattern speed information from driving logs.
While many recent approaches have primarily focused on capturing relationships between neighbor links and congestion propagation based on real-time speed information, they often employ complex methodologies such as representing the road network as a graph, applying GNNs with multi-step operations, or predefining strongly correlated distant links.
In contrast, our proposed model focuses on the overall driving pattern across the road network and effectively leverages it for ETA prediction.
Although individual driving instances are initiated by diverse and often unpredictable personal factors, traffic flow, when observed collectively across a large population and over time, tends to exhibit consistent and recurring patterns.
These patterns arise from the aggregation of countless individual driving behaviors, which are themselves influenced by both global and local factors. Global factors—such as time of day, day of week, and broader social routines—contribute to statistically stable spatiotemporal distributions of traffic speed across road segments. In contrast, local factors—including regional events, construction, and accidents—are irregular in nature and, while they can have strong localized impacts on traffic, their influence on the overall traffic pattern tends to be limited.
Furthermore, if such local disruptions have already occurred, their effects are often reflected in the current states of adjacent links. For distant disruptions that have not yet propagated, the impact is typically minor, resulting in a limited effect on the broader traffic pattern.

Based on this insight, if sufficient pattern information and a reliable representation of the current state are provided, it is possible to achieve high ETA prediction accuracy without explicitly modeling complex relationships between neighbor links.
Based on this perspective, our model focuses on the use of pattern speed and the inherent characteristic of the ETA task itself rather than on representing the relationships between neighbor links.
To utilize pattern speed effectively, we apply an attention mechanism over temporal slots of the pattern speed based on temporal embedding vectors and environmental features.
Since the ETA task inherently involves traversing a sequence of road links, the driving behavior on the current link is causally dependent only on the previously traversed links, making it a spatio-temporal causal problem. This spatio-temporal causality has led prior methods to adopt sequential link-wise duration prediction, rather than simultaneous estimation across all links.

To address this, we propose a novel method that accumulates and passes a time pre-projection vector along the route, combined with k-cycle processing that allows the model to refine its predictions iteratively.
Typically, the entry time into a given link is represented as a scalar sum—the sum of durations of previous links. However, our method encodes this time information as a vector, enabling the model to convey richer temporal representations while also capturing uncertainty arising during accumulation.
This vectorized temporal embedding allows for more informed and flexible ETA predictions, and repeated refinements over multiple cycles further enhance prediction accuracy and robustness.

Consequently, our model demonstrates substantially improved prediction performance and robustness, offering a practical and efficient solution for real-world ETA estimation tasks.

\section{Related Works}

To address this challenge, a wide variety of ETA prediction methods have been developed within the field of Intelligent Transportation Systems (ITS).
One of the earlier approaches, Similarity-based travel time estimation \cite{sbtte}, introduced a lightweight model that identifies historically similar trips from origin-destination trip pairs and estimates ETA by adjusting historical speed data through scaling factors. This approach accounts for both temporal dynamics and geographic regional characteristics, and it offers the advantage of computational efficiency. However, its reliance on historical pattern matching limits its generalizability to rare or unseen routes and significantly diminishes its ability to adapt to real-time conditions.
To address these limitations, Sparse trajectory based travel time estimation \cite{ttepst} proposed a more structured modeling framework. This approach constructs a large sparse 3D tensor incorporating road segments, time slots, and driver identifiers. The tensor is then utilized to extract frequent trajectory patterns, decompose travel routes into relevant sub-paths, and aggregate time costs across these sub-paths. This structured representation enables more informed and interpretable ETA predictions.

With recent advancements in deep learning, many neural network-based approaches have been proposed, offering substantial improvements over traditional methods.
For instance, DeepTravel \cite{deeptravel} avoids the use of explicit road network graphs by discretizing geographic space into a two-dimensional N × N grid. It models travel sequences in three distinct stages—starting, middle, and ending—while leveraging both short-term and long-term traffic features to predict travel time.
DCRNN \cite{dcrnn} improves the modeling of spatiotemporal dependencies by combining diffusion convolution, applied over directed graphs representing the road network, with GRU\cite{gru}-based temporal sequence modeling.
Similarly, ST-GCN \cite{stgcn} views traffic data as a graph-structured time series and leverages spatial graph convolutions with temporal gated convolutions. This design allows the model to capture spatiotemporal relationships effectively while avoiding the computational complexity associated with recurrent layers.
STANN \cite{stann} further enhances spatiotemporal modeling by introducing an encoder-decoder architecture with both spatial and temporal attention mechanisms, enabling more accurate long-range traffic predictions.

More recently, GNNs \cite{gnn1, gnn2} have emerged as a powerful modeling paradigm for ETA prediction, particularly due to their capacity to capture the topological structures and complex dynamics of road networks.
CompactETA \cite{compacteta} proposes an architecture consisting of an asynchronous updater and a lightweight predictor. The updater employs a graph attention network that runs recurrently for $k$ steps to generate a pre-computed table of link representations, capturing multi-hop neighbor influences. During inference, the predictor retrieves relevant representations from this table using a lightweight query mechanism enhanced by positional encoding, allowing CompactETA to effectively model sequential travel while maintaining computational efficiency.
GNN-ETA \cite{gnneta} presents a large-scale GNN model designed for real-world production environments. It estimates travel times over supersegments by integrating both real-time and historical traffic features, and it employs stabilization techniques such as MetaGradients \cite{metagrad} and Exponential Moving Averages (EMA) to ensure robustness in real-world production environments.
ConSTGAT \cite{constgat} refines GNN-based modeling by introducing a three-dimensional graph attention network that jointly captures spatial and temporal correlations. Instead of treating spatial and temporal dynamics separately, ConSTGAT builds a 3D graph where nodes correspond to road segments at different time slots, allowing attention to be applied across both space and time simultaneously. After the attention-based aggregation, convolutional windows are used to summarize local contextual information, which is then leveraged to predict travel times at both the segment and route levels.
Meanwhile, DuETA \cite{dueta} introduces a congestion-aware graph transformer architecture that jointly models traffic congestion propagation and long-range dependencies across the network. By leveraging both a congestion-sensitive graph and a route-aware transformer, DuETA captures not only local interactions but also distant correlations among traffic segments. This design allows the model to remain robust under complex conditions, such as cascading traffic delays, which are challenging for conventional approaches to handle.

\section{Method}

\subsection{Problem Formulation}

We formulate ETA prediction as a supervised regression task, where the goal is to estimate the total duration of a given vehicle trajectory based on background information, road context data, and historical speed patterns.
A vehicle route is composed of the background information and a sequence of connected links, denoted by:
\begin{align}
    (\mathbf{R}, \mathbf{b}) = \left([\mathbf{r}_1, \mathbf{r}_2, \dots, \mathbf{r}_n], \mathbf{b}\right)
\end{align}
where $\mathbf{b}$ represents global environmental factors, such as the day of the week and the departure time, which significantly affect traffic patterns, and $\mathbf{R}$ denotes the ordered sequence of links that constitute the route.

Each link $\mathbf{r}_i$ is associated with a set of features consisting of:
\begin{align}
    \mathbf{r}_i = (\mathbf{c}_i, \mathbf{p}_i)
\end{align}
where $\mathbf{c}_i$ represents the static and dynamic attributes of the link (e.g., road kind, number of lanes, curvature, real-time speed at departure), and $\mathbf{p}_i$ denotes the pattern speed profile, which consists of historical average speeds aligned by day of week and time of day.

The total travel duration $d$ is defined as the sum of durations over all links:
\begin{align}
    D = \sum_{i=1}^{n} d_i
\end{align}
Each link duration $d_i$ is modeled as a function of the background information $\mathbf{b}$, the link features $(\mathbf{c}_i, \mathbf{p}_i)$, and the link entry time $t_i$:
\begin{align}
    d_i = f_{\mathrm{link}}(\mathbf{b}, \mathbf{c}_i, \mathbf{p}_i, t_i)
\end{align}
where the link entry time $t_i$ is computed recursively based on the cumulative durations of the preceding links:
\begin{align}
    t_i = t_0 + \sum_{k=1}^{i-1} d_k, \quad \text{with} \quad t_0 = \text{departure time}
\end{align}
The objective is to learn a function $f$ that captures these dynamic interactions and outputs the estimated total duration:
\begin{align}
    \hat{D} = f(\mathbf{b}, \mathbf{R}, t_0)
\end{align}
where $\hat{D} \in \mathbb{R}^+$ is a scalar prediction representing the estimated travel time.

This formulation enables the model to account for the temporal shift of traffic conditions along the route and to leverage both local road-level patterns and global temporal context for accurate ETA prediction.

\subsection{Structure of PAtt model}

\begin{figure}[tbp]
\centering{
\includegraphics[width=1.0\linewidth]{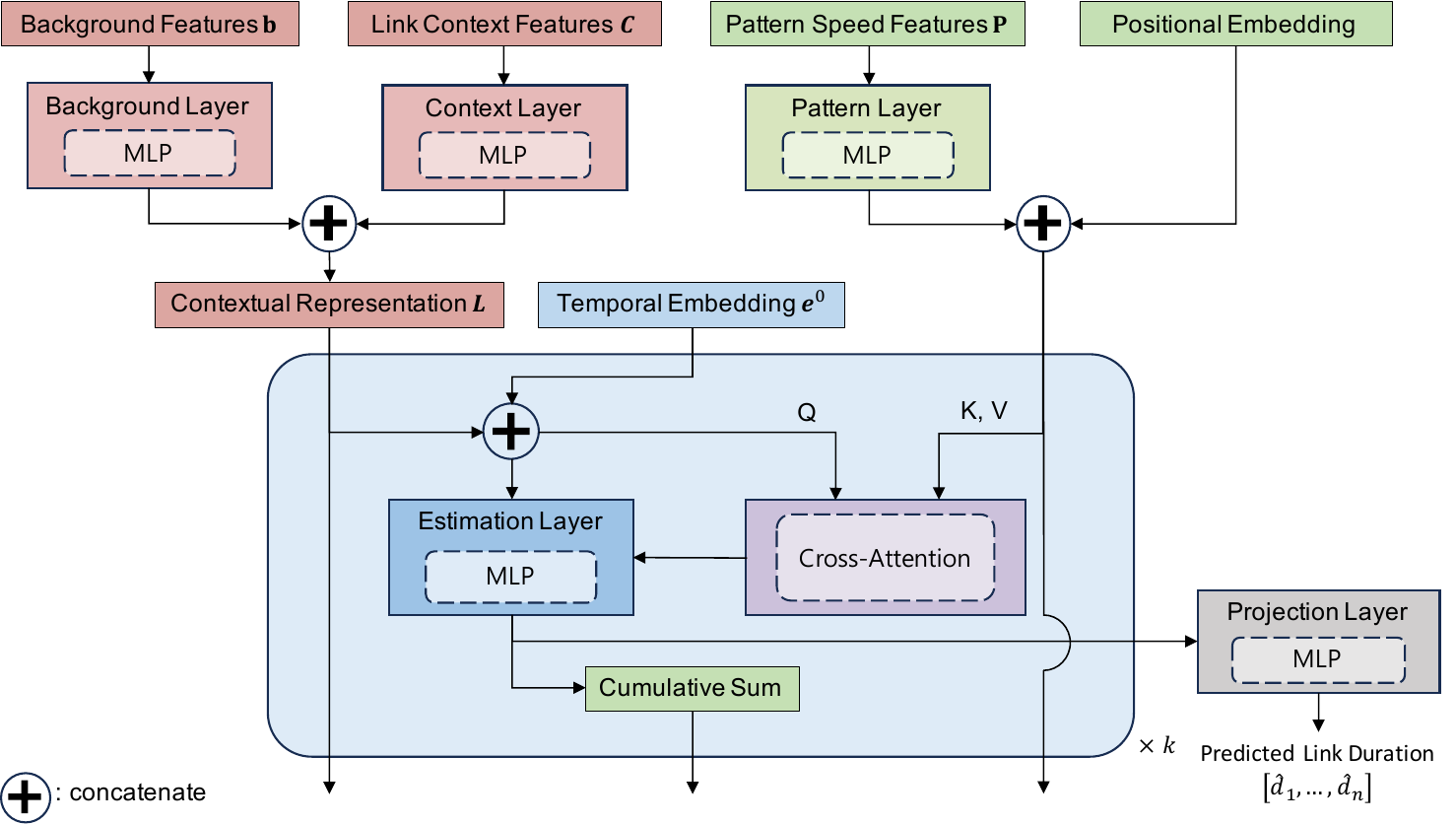}
}
\caption {{\bf The overall architecture of Model:} Contextual representation are extracted from background, link context via MLP layers, combined with temporal embedding, and processed through a cross attention with pattern speed profile and an estimation layer to produce temporal embedding. The procedure is repeated k times and the final temporal embedding is used to produce the final link-level ETA predictions.}
\label{fig:all_system}
\end{figure}

\subsubsection{Overall System}
Fig. \ref{fig:all_system} illustrates the overall structure of the proposed PAtt (Pattern Attention) model.
The model processes input features by decomposing them into three components: (1) a global background feature vector $\mathbf{b}$, (2) per-link context feature vectors $\mathbf{C} = [\mathbf{c}_1, \mathbf{c}_2, \dots, \mathbf{c}_n]$, and (3) per-link pattern speed profile vectors $\mathbf{P} = [\mathbf{p}_1, \mathbf{p}_2, \dots, \mathbf{p}_n]$ defined over temporal slots.

At the initial cycle, the temporal embedding vectors are initialized as zero vectors. The background feature vector $\mathbf{b}$, which remains constant across all links, is provided to each link-level processing unit. Simultaneously, each link context feature $\mathbf{c}_i$ is combined with $\mathbf{b}$ to form a contextual representation $\mathbf{l}_i$.

For each link, a pattern attention mechanism is applied. The contextual representations $\mathbf{L} = [\mathbf{l}_1, \mathbf{l}_2, \dots, \mathbf{l}_n]$ is used as the query, while the pattern speed profiles $\mathbf{P}$—a temporal sequence of historical average speeds—are used as both the keys and values. The outputs of the attention module are concatenated with the corresponding contextual representations to produce the temporal embedding vectors for each link.

These embeddings are intermediate representations, not direct duration estimates.
At each cycle, the embeddings are cumulatively updated along the route and then used as inputs for the next cycle.
As the model proceeds through $k$ refinement cycles, the temporal embeddings for each link are progressively refined.

After $k$ refinement cycles, the final temporal embedding vectors are projected into scalar values representing the estimated duration of each link.
The total estimated duration is computed by summing all per-link predictions.

\subsubsection{Feature Representation}
The input features of the PAtt model consist of three main components: background features, link context features, and pattern speed features. Each component captures different aspects of temporal and spatial traffic information required for accurate duration prediction.
Raw feature values from different components are first preprocessed and then individually transformed through separate MLP layers to produce dense feature vectors suitable for downstream processing. This design allows the model to flexibly handle heterogeneous feature types and to learn appropriate representations for each component.

\paragraph{Background Features ($\mathbf{b}$)}
Global features that are constant across all links within a trip. These include:
\begin{itemize}
    \setlength{\itemindent}{2em}
    \item Day of the week
    \item Time of departure (e.g., hour or minute of day)
    \item Holiday indicator
\end{itemize}

\paragraph{Link Context Features ($\mathbf{C}$)}
Per-link features describing the physical, dynamic, and traffic-related characteristics of each link. These include:
\begin{itemize}
    \setlength{\itemindent}{2em}
    \item Static attributes: road kind, number of lanes, curvature, link length, speed limit
    \item Real-time speed at the time of departure
    \item Representative average speed
\end{itemize}

\paragraph{Pattern Speed Features ($\mathbf{P}$)}
A sequence of historical average speeds for each link, retrieved according to future time slots relative to the departure time (e.g., every 20 minutes up to a 120-minute horizon). These historical profiles represent typical traffic conditions at future moments from the departure time and capture global and periodic traffic flow patterns. Formally, for link $\mathbf{r}_i$, we define:
\begin{align}
\mathbf{p}_i = [p_i^{(1)}, p_i^{(2)}, \dots, p_i^{(T)}], \quad p_i^{(t)} \in \mathbb{R}    
\end{align}
where $T$ denotes the number of future time slots.

\paragraph{Feature Normalization and Alignment}
Numerical features are normalized using either min-max scaling or z-score standardization, and categorical features are embedded via learnable embeddings. Speed-related features are further processed through min-max normalization within predefined bounds, quantization, one-hot encoding, and optional embedding. All features are aligned to ensure consistency across time and the sequential order of links.

\subsubsection{Pattern Attention Network}
The Pattern Attention Network is designed to extract temporally-aware link representations by leveraging pattern speed profiles through an attention mechanism. It operates in $k$ iterative refinement cycles, where each cycle updates the temporal representation of each link using aggregated contextual and pattern-based information.

\paragraph{Extract Link Representation}
Initially, the background feature vector $\mathbf{b}$ and the link context feature $\mathbf{C}$ are concatenated and passed through a multilayer perceptron (MLP) to generate a high-level link representation $\mathbf{L}$:
\begin{align}
\mathbf{L} = \mathrm{MLP}_{\mathrm{bc}}([\mathbf{b}; \mathbf{C}])
\end{align}
In the $m$-th cycle, this representation is concatenated with the pre-projection temporal embedding vector $\mathbf{e}_i^{(m-1)}$ from the previous cycle (initialized as zero in the first cycle), forming the query vector $\mathbf{q}_i^{(m)} = [\mathbf{l}_i; \mathbf{e}_i^{(m-1)}]$ for attention.

\paragraph{Pattern Attention over Temporal Slots}
A soft attention mechanism is applied over these temporal slots:
\begin{align}
\mathbf{a}_i^{(m)} = \mathrm{Attention}(\mathbf{q}_i^{(m)}, \mathbf{k}_i, \mathbf{v}_i), \quad \text{where } \mathbf{k}_i = \mathbf{v}_i = \mathbf{p}_i
\end{align}
The output $\mathbf{a}_i^{(m)}$ encodes how the temporal dynamics relate to the current query.
Through this mechanism, the model learns to attend to time slots that are most relevant to the current traffic context.

\paragraph{Temporal Embedding Refinement}
The attention output $\mathbf{a}_i^{(m)}$ is concatenated with the link representation $\mathbf{l}_i$ and passed through another MLP to generate the updated pre-projection temporal embedding:
\begin{align}
\mathbf{e}_{i,link}^{(m)} = \mathrm{MLP}_{\mathrm{emb}}([\mathbf{a}_i^{(m)}; \mathbf{l}_i])
\end{align}
After generating $\mathbf{e}_{i,link}^{(m)}$ for all links, they are cumulatively summed over previous links up to $i-1$ to reflect the temporal shift in traffic:
\begin{align}
\mathbf{e}_i^{(m)} = \Sigma_{j=1}^{i-1}\mathbf{e}_{j,link}^{(m)}
\end{align}
This updated embedding is then used again as part of the next cycle’s query input.

\paragraph{Final Projection}
After $k$ refinement cycles, the final temporal embeddings $\mathbf{e}_{i,link}^{(k)}$ are passed through a projection MLP to produce the predicted duration $\hat{d}_i$ for each link:
\begin{align}
\hat{d}_i = \mathrm{MLP}_{\mathrm{proj}}(\mathbf{e}_{i,link}^{(k)})
\end{align}
The total travel time $\hat{D}$ is obtained by summing all per-link predictions:
\begin{align}
\hat{D} = \sum_{i=1}^{n} \hat{d}_i
\end{align}

\subsection{Loss Function}
To train the PAtt model, we define a composite loss that accounts for both fine-grained link-level accuracy and global route-level consistency. The overall loss is composed of two components: the link-level loss and the route-level loss.
\paragraph{Link-Level Loss} For each link $r_i$, the model predicts the travel duration $\hat{d}_i$, which is compared to the ground-truth duration $d_i$. We use the Smooth L1 loss (also known as Huber loss) to provide robustness against outliers while maintaining sensitivity to small errors:
\begin{align}
\mathcal{L}_{\text{link}} = \frac{1}{n} \sum_{i=1}^{n} \mathrm{SmoothL1}(\hat{d}_i, d_i)
\end{align}
This loss encourages accurate per-link predictions and facilitates stable training by controlling the gradient scale.
\paragraph{Route-Level Loss}
To ensure that the overall route-level prediction aligns with the true total travel duration, we define an accumulation on the total predicted duration $\hat{D}$ compared to the ground-truth duration $D$.
We adopt the Absolute Percentage Error (APE) as the route-level loss, while introducing a small constant $D_{low} = 20$ (minutes) to stabilize the denominator and prevent instability caused by disproportionately large errors on short routes.
\begin{align}
    \mathcal{L}_{\text{route}} = \left| \frac{\hat{D} - D}{\max(D, D_{low})} \right|
\end{align}
Here, $D_{low}$ acts as a lower bound for the denominator, ensuring that the percentage error remains well-behaved even when $D$ is small.
\paragraph{Multi-Cycle Aggregation}
To encourage consistent refinement across all cycles, both losses are computed not only on the final outputs but also on the intermediate outputs from each refinement cycle $m = 1, \dots, k$. Let $\hat{d}_i^{(m)}$ denote the predicted duration for link $\mathbf{r}_i$ at cycle $m$, and $\hat{D}^{(m)} = \sum_{i=1}^{n} \hat{d}_i^{(m)}$ be the corresponding total route duration. The overall training loss is the sum of all link and route losses over $k$ cycles:
\begin{align}
\mathcal{L}_{\text{total}} = \sum_{m=1}^{k} \left(\lambda \cdot \mathcal{L}_{\text{link}}^{(m)} + \mathcal{L}_{\text{route}}^{(m)} \right)
\end{align}
This multi-cycle supervision strategy stabilizes learning and encourages progressive refinement of predictions throughout the pattern attention iterations.

\section{Experiments}

\subsection{Experimental Setup}

\subsubsection{Dataset}

We use a large-scale real-world dataset collected from South Korea for training and evaluation.
The dataset includes vehicle trajectories annotated with per-link travel durations, as well as dynamic traffic features such as pattern speed profiles. The data spans the period from March 1st, 2024, to February 28th, 2025.
To ensure the robustness of the model, we filter out extremely short or long routes and retain only those with lengths between 5 km and 500 km. This filtering results in a total of 294,186,507 valid routes. We use 90\% of these samples for training and the remaining 10\% for evaluation. On average, each route contains 47 links, with a mean route length of 28 km and an average travel duration of 35 minutes.

\subsubsection{Models for Comparison}

We compare the proposed model against the following baselines and representative methods:

\paragraph{Rule-Based Speed Combination (Baseline)}
This baseline computes ETA by combining real-time and pattern-based speed profiles. For each link, a weighted average speed is calculated, where the weights are determined based on the expected entry time into the link.

\paragraph{DeepTravel \cite{deeptravel}}
DeepTravel is an end-to-end ETA prediction model that transforms routes into grid sequences. It employs a bidirectional LSTM with dual auxiliary losses to learn both forward and backward cumulative travel times. This design enhances the model's ability to generalize across variable-length routes and facilitates robust learning of temporal travel patterns.

\paragraph{CompactETA \cite{compacteta}}
CompactETA generates spatio-temporal link representations using a graph attention network. It then utilizes positional encoding to retain the sequential context of routes without relying on recurrent operations, instead using precomputed link representations for efficient inference.

We exclude several recent models from direct comparison for the following reasons:
\begin{itemize}
    \item GNN-ETA \cite{gnneta}, DCRNN \cite{dcrnn} and STANN \cite{stann} predict ETA at the segment or link level and evaluate performance accordingly, which is not directly comparable with our route-level prediction setting.
    \item DuETA is particularly designed for short-range urban driving scenarios and is not suited for evaluating models over the long and diverse routes considered in our experiments.
\end{itemize}

\subsection{Experiment Results}

\subsubsection{Ablation Study}

\begin{figure}[tbp]
    \centering
    \subfigure[MAE for each loss function by number of cycles]{
        \centering
        \includegraphics[width=0.9\linewidth]{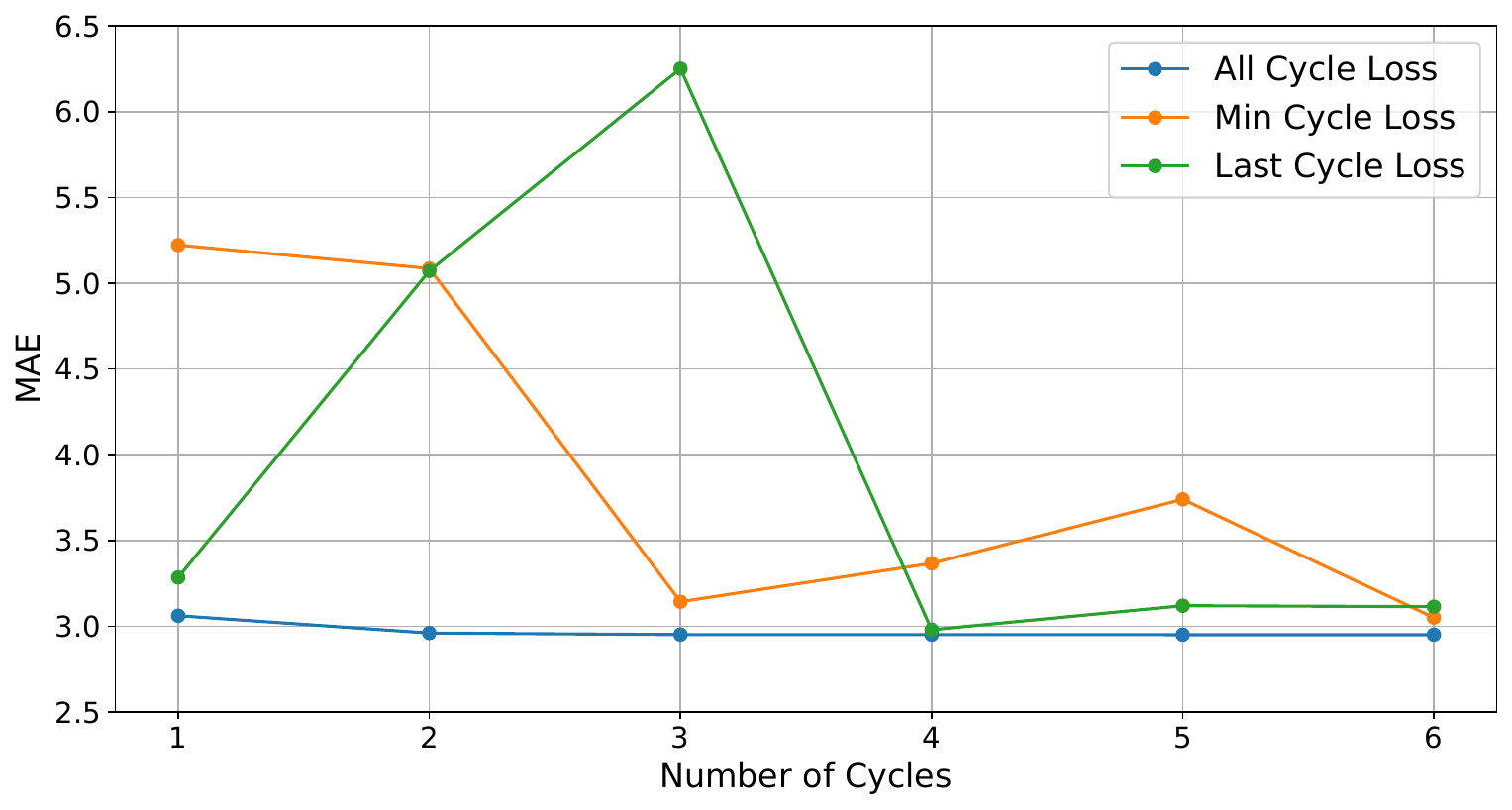}
        % \caption{MAE for each loss function by number of cycles}
    }
    \hfill
    \subfigure[MAPE for each loss function by number of cycles]{
        \centering
        \includegraphics[width=0.9\linewidth]{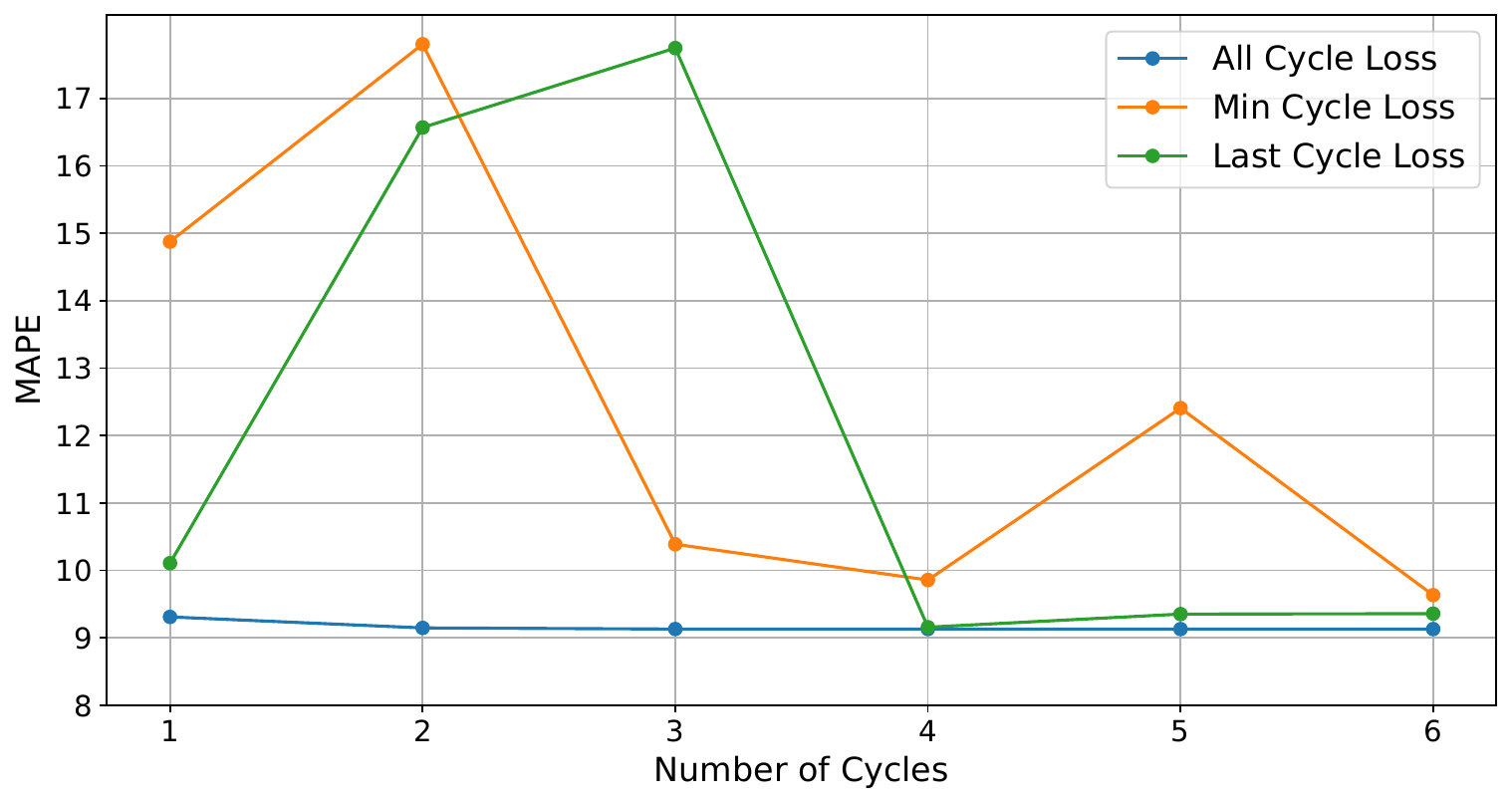}
        % \caption{MAPE for each loss function by number of cycles}
    }
    \caption{Performance comparison of loss functions with varying number of cycles}
    \label{fig:loss_cycle}
\end{figure}

In this section, we evaluate how different loss functions and the number of refinement cycles affect training and performance, as illustrated in Figure~\ref{fig:loss_cycle}.
We compare three training strategies:
\begin{itemize}
    \item All Cycle Loss: computes the total loss across all refinement cycles
    \item Min Cycle Loss: uses the lowest loss among all cycles for training
    \item Last Cycle Loss: uses only the last cycle's loss during training
\end{itemize}
For all experiments, the number of cycles for training was set to 4.
As shown in Figure~\ref{fig:loss_cycle}, the All Cycle Loss consistently outperforms the other loss strategies across all cycles. While Min Cycle Loss and Last Cycle Loss exhibit unstable performance, with fluctuations in both MAE and MAPE without clear trends as the number of cycles increases, All Cycle Loss demonstrates stable and gradually improving results. This confirms the effectiveness of iterative refinement and supports our design choice of leveraging all cycles for training.

\subsubsection{Quantitative Result}

\paragraph{Performance Comparison}

We compare our model against several baseline and recent methods in terms of Mean Absolute Percentage Error (MAPE). While MAE and RMSE are also commonly used, they are highly sensitive to the distribution characteristics of the dataset, making them less suitable for cross-dataset comparisons.

\begin{table}[tbp]
\caption{Comparison of MAPE across the proposed method and existing ETA prediction models}
\begin{center}
\begin{tabular}{|c|c|c|}
\hline
Method   & MAPE(\%) \\ \hline
Baseline & 12.1  \\
DeepTravel\textsuperscript{*} &  13.30   \\
CompactETA\textsuperscript{*} & 11.07  \\
DuETA\textsuperscript{*} &  19.50 \\
PAtt  &   8.78   \\
\hline
\multicolumn{2}{r}{\textsuperscript{*} Results are quoted from the corresponding references.} \\
\end{tabular}
\label{table:comp}
\end{center}
\end{table}

As shown in Table~\ref{table:comp}, our proposed model achieves the lowest MAPE, significantly outperforming both the traditional baseline and recent deep learning-based models.

\paragraph{Performance Across Route Lengths}

\begin{figure}[tbp]
    \centering
    \subfigure[MAE according to route length]{
        \centering
        \includegraphics[width=0.9\linewidth]{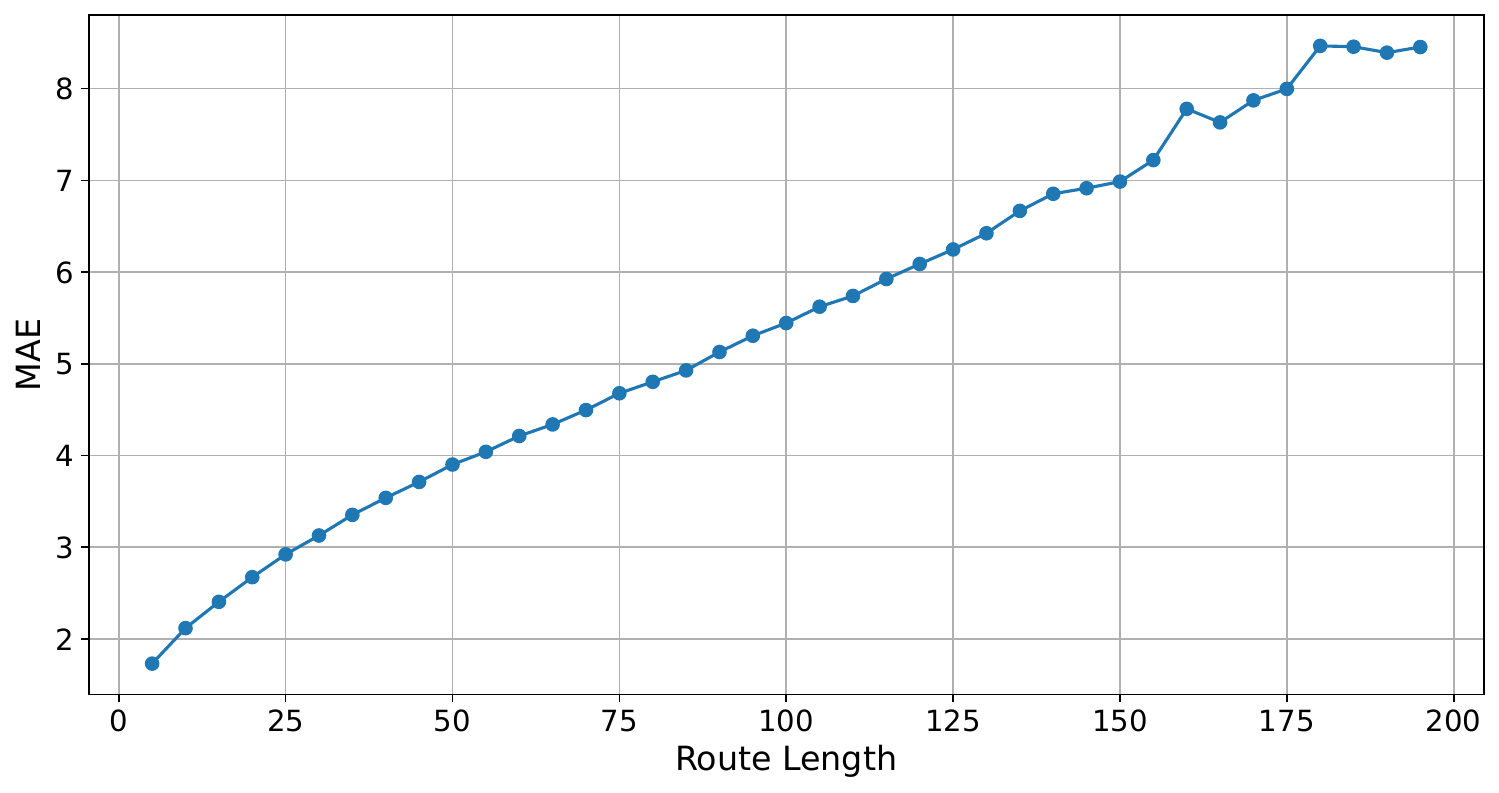}
        % \caption{MAE according to route length}
        \label{fig:route_perform_mae}
    }
    \hfill
    \subfigure[MAPE according to route length]{
        \centering
        \includegraphics[width=0.9\linewidth]{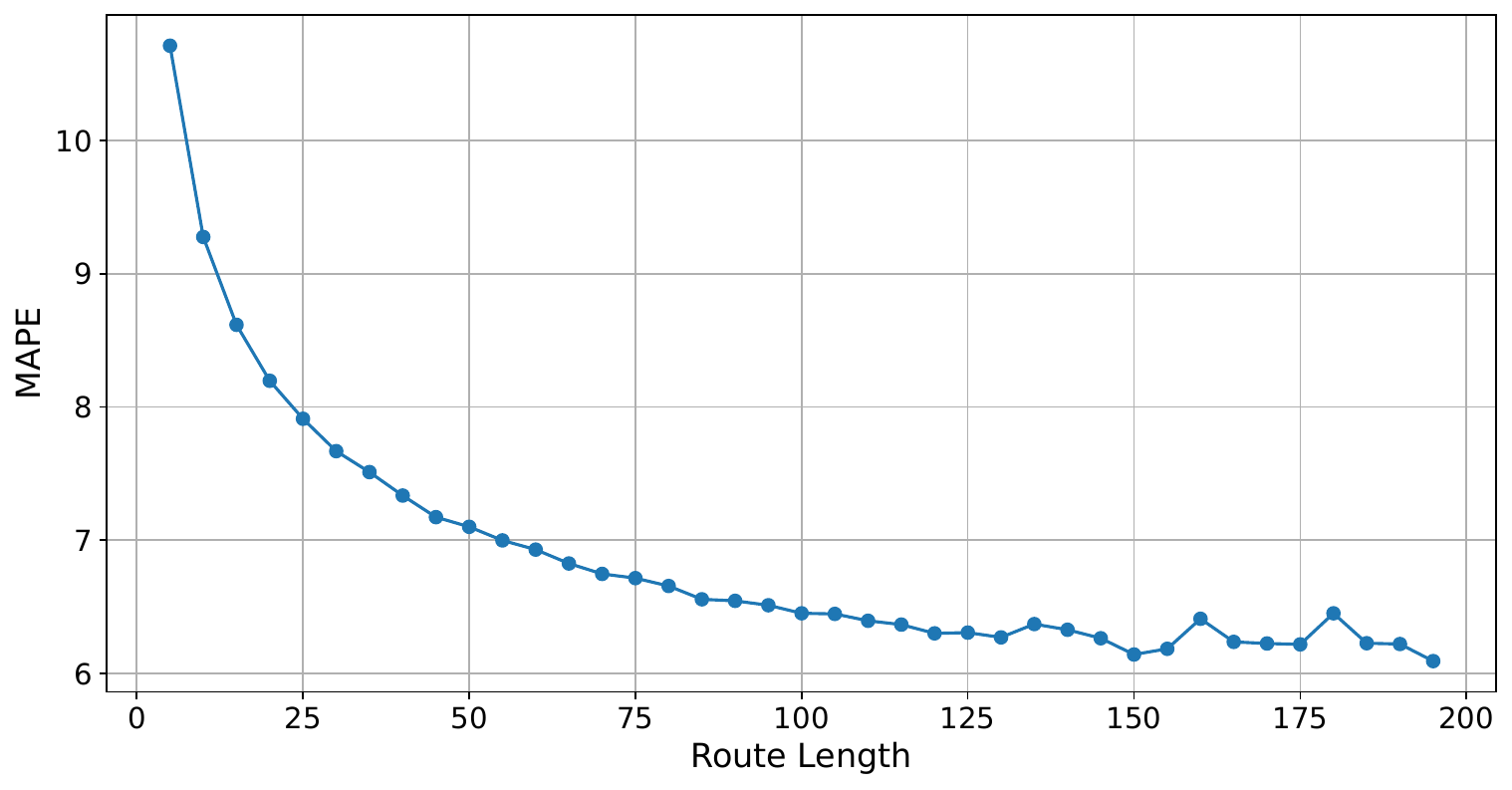}
        % \caption{MAPE according to route length}
        \label{fig:route_perform_mape}
    }
    \caption{Performance variation with respect to route length}
    \label{fig:route_perform}
\end{figure}

We further analyze how the model performance varies according to route length. Figure~\ref{fig:route_perform} presents the MAE and MAPE of the proposed model across different route lengths.
As shown in Figure~\ref{fig:route_perform_mae}, the MAE increases with route length, which is an expected outcome since absolute travel time errors tend to grow proportionally with longer distances and durations.
However, Figure~\ref{fig:route_perform_mape} shows that the MAPE decreases as route length increases, indicating that the model achieves relatively more accurate predictions for longer routes when considering proportional errors.
This trend demonstrates that the proposed model effectively leverages historical pattern speed profiles to forecast future traffic states even over longer time horizons.
In addition, Figure~\ref{fig:route_perform_mape} shows that the model maintains robust performance even for short-distance routes, achieving a MAPE consistently below 11\%. 
Although short routes are inherently more sensitive to small absolute errors (leading to higher MAPE), the proposed method still delivers reliable results. These findings demonstrate that our model not only handles long-range predictions effectively but also preserves strong predictive performance across diverse route lengths.

\section{Conclusions} \label{conclusions}

In this paper, we proposed a novel ETA prediction model that effectively leverages historical pattern speed profiles through a Pattern Attention mechanism. Unlike existing approaches that heavily rely on capturing real-time neighbor link interactions or complex road network structures, our method focuses on extracting temporally-aware representations based on global and local traffic patterns.
Through extensive experiments on large-scale real-world datasets, we demonstrated that the proposed model achieves superior performance compared to baselines, particularly in handling both short and long routes. The results confirm that incorporating structured pattern information not only improves predictive accuracy but also enhances model robustness across various traffic conditions.
Our findings highlight the importance of historical traffic patterns in ETA prediction and suggest that lightweight, pattern-driven architectures can serve as strong alternatives to more complex graph-based models.
In future work, we plan to further explore adaptive pattern extraction techniques and investigate the integration of real-time local event information to enhance model responsiveness.

{\small 
\bibliographystyle{ieeetr}
\bibliography{egbib}
}

\end{document}